\documentclass{article}
\usepackage{spconf,amsmath,amsfonts,epsfig}
\usepackage{array}
\usepackage{caption}
\usepackage{textcomp}
\usepackage[dvipsnames]{xcolor}
\usepackage{graphicx}
\usepackage[colorlinks=true,linkcolor=NavyBlue,citecolor=BrickRed]{hyperref}
\usepackage{cleveref}
\usepackage{subfigure}
\newcommand{\CRediT}{\textsuperscript{\small CRediT}}
\newcommand{\DeepLCZChange}{\textit{DeepLCZChange}}
\title{\DeepLCZChange:\\A Remote Sensing Deep Learning Model Architecture \\ for Urban Climate Resilience\CRediT}
\name{Wenlu Sun$^1$, Yao Sun$^1$, Chenying Liu$^{1,2}$, Conrad M Albrecht$^2$
\thanks{\CRediT~\textit{Contributor Role Taxonomy} statement, \url{https://credit.niso.org} ---
\textbf{Wenlu Sun}: Data curation, Software, Visualization, Writing--original draft;
\textbf{Conrad M Albrecht}: Conceptualization, Methodology, Formal Analysis, Software, Visualization, Resources, Writing--original draft, Writing--review \& editing, Supervision;
\textbf{Chenying Liu}: Software, Data curation, Writing--review \& editing;
\textbf{Yao Sun}: Writing--review \& editing, Supervision;
}
\thanks{Corresponding author: \texttt{Conrad.Albrecht@DLR.de}}
\address{
$^1$ Data Science in Earth Observation, Technical University of Munich\\
$^2$ Remote Sensing Technology Institute, German Aerospace Center
}}
\begin{document}
%\ninept
%
\maketitle
\begin{abstract}
Urban land use structures impact local climate conditions of metropolitan areas. To shed light on the mechanism of local climate wrt.\ urban land use, we present a novel, data-driven deep learning architecture and pipeline, \textit{DeepLCZChange}, to correlate airborne LiDAR data statistics with the Landsat 8 satellite's surface temperature product. A proof-of-concept numerical experiment utilizes corresponding remote sensing data for the city of New York to verify the cooling effect of urban forests.%
%In contrast to regular deep learning neural network optimization, we consider fixed model parameters, but variation of model input data, i.e. LiDAR statistics.
\end{abstract}
\section{Introduction} %\& Motivation}
\label{sec:intro}
\begin{figure}[t!]
\centering
\includegraphics[width=.53\textwidth]{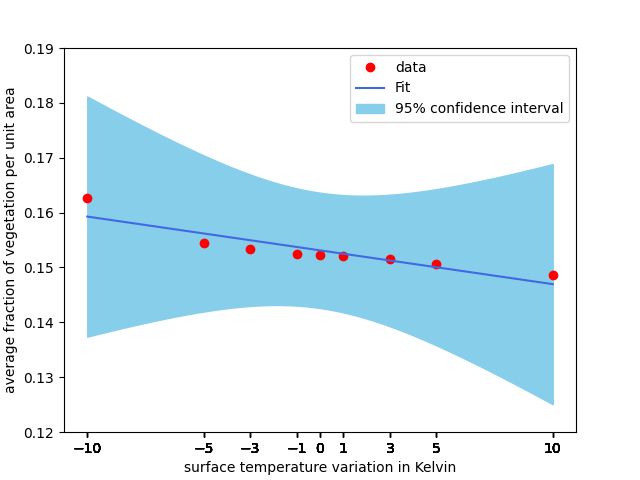}
\caption{Data-driven correlation analysis of urban surface temperature vs.\ vegetation cover from satellite-based thermal measurements vs.\ airborne LiDAR point cloud statistics. A linear fit (blue line) reveals a negative slope of $-6\cdot10^{-4}/K$ with $R^2=.82$ from data (red) generated by the \textit{DeepLCZChange} artificial neural network.}
\label{fig:result}
\end{figure}

Projections estimate about 2/3 of the global population is going to cluster in urban spaces by the end of 2030~\cite{WorldUrbanization2018}. \textit{Urban land use structures} (ULUS) have an impact on local climate conditions~\cite{SALAZAR2015103}, e.g.\ reduced airflow may slow down heat dispersion inducing negative consequences for local ecosystems through \textit{surface urban heat islands} \cite{SUHI_Intro}. Shedding light onto the interaction of \textit{local climate zones} (LCZ)~\cite{stewart2012local} and corresponding climate proxies is a key for urban planning towards climate resilience.

\begin{figure*}[h!]{
  \centering\hfill
  \subfigure{\includegraphics[width=0.24\textwidth]{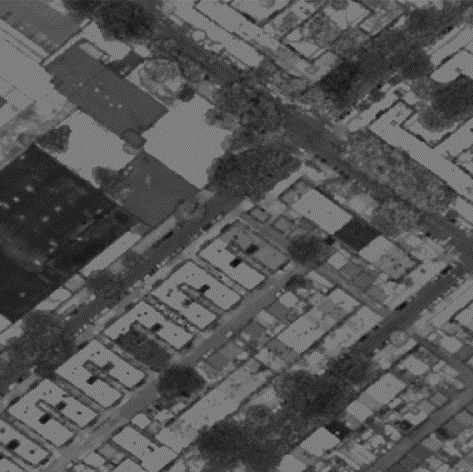}\label{fig:laserreflect}}\hfill
  \subfigure{\includegraphics[width=0.24\textwidth]{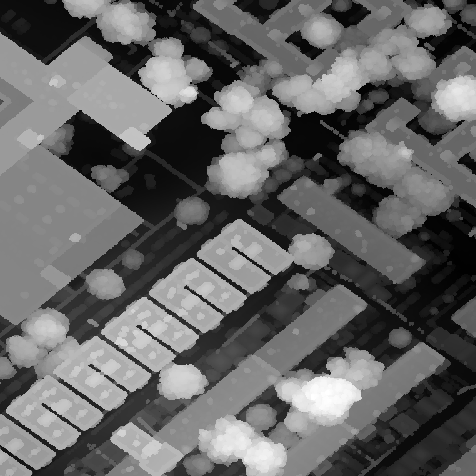} \label{fig:triangle}}\hfill
  \subfigure{\includegraphics[width=0.24\textwidth]{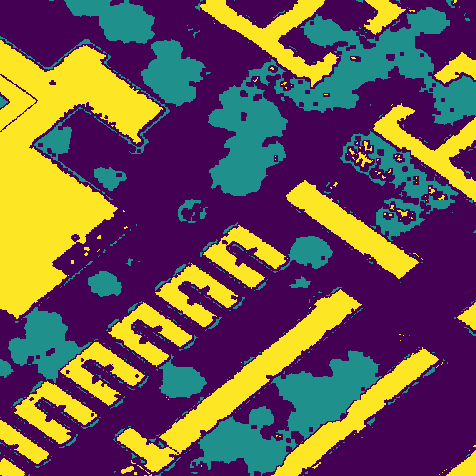}\label{fig:treebuildingmap}}\hfill\hfill
  \caption{\textit{Visualization of urban scene}: Top-down view derived from three-dimensional LiDAR point cloud as sensed by laser reflectance (left) and laser pulse time-of-flight, i.e. elevation map (center). The semantic segmentation (right) is based on the \textit{AutoGeoLabel} procedure, cf.\ post-processing in \Cref{sec:experiment}, with yellow buildings, green vegetation, and dark purple background.}
  \label{fig:three}       }
\end{figure*}

Our data-driven approach detailed in \Cref{sec:method} studies correlations of ULUS and \textit{Local Surface Temperature} (LST) to explore how urban development impacts ambient temperatures, cf.\ \Cref{sec:experiment}. Technically, we utilize statistics of an airborne LiDAR survey in New York City\footnote{\url{https://maps.nyc.gov/lidar/2017}} which bears semantic signatures of, e.g., vegetation and buildings~\cite{albrecht2022monitoring}, cf.\ \Cref{fig:three} for illustration. We co-register the LiDAR statistics with the surface temperature product of the Landsat 8 satellite \cite{cook2014development}. Based on those data, we  propose a novel deep neural network architecture termed \textit{DeepLCZChange} to model correlations between vegetation and ambient temperature.

\section{Methodology}
\label{sec:method}
\begin{figure}[t!]
\centering
\includegraphics[width=.4\textwidth]{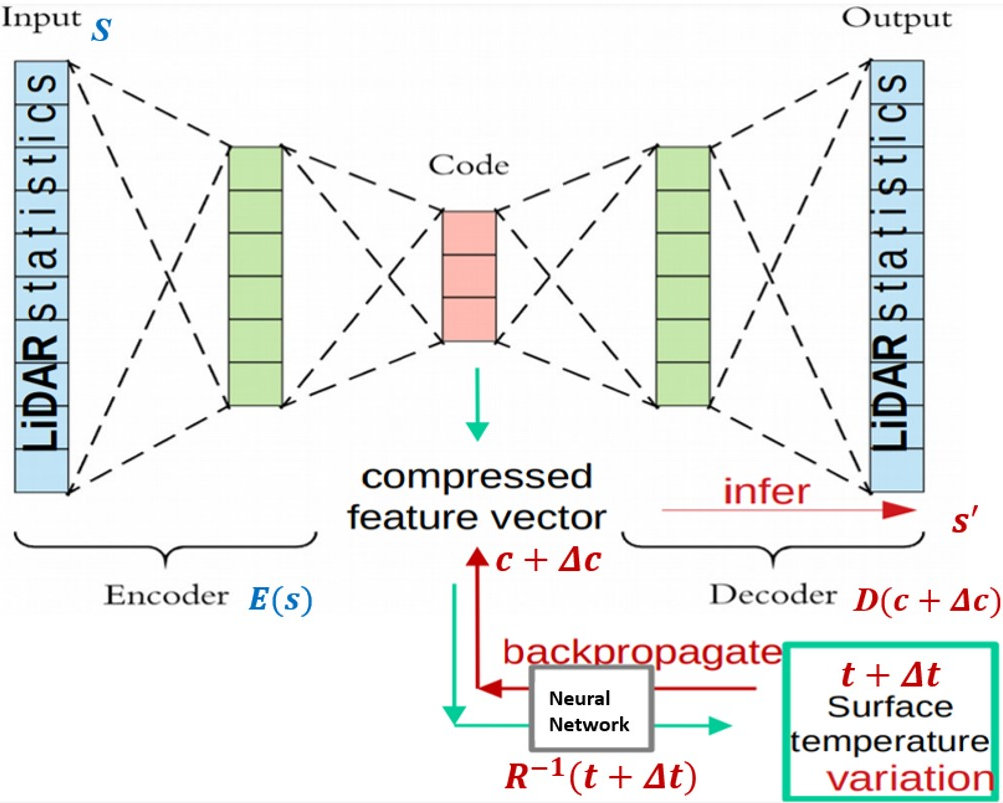}
\caption{\textit{DeepLCZChange}: Artificial neural network architecture and three-stage methodology: 1.\ unsupervised autoencoder training (on high resolution urban remote sensing modality), blue; 2.\ supervised downstream task (on low resolution satellite data as climate proxy), green; and 3.\ backpropagation as inference step for generative modelling (correlating autoencoding of 1.\ with \textit{downstream tasking} of 2.), red.}
\label{fig:architecture}
\end{figure}

\textit{DeepLCZChange} introduces a three-stage deep learning procedure summarized by \Cref{fig:architecture}. In a first step a \textit{variational autoencoder} (VAE), $D\circ E$, compresses the LiDAR statistics $s$ into representations $c=E(s)$, roughly speaking by minimizing the loss $\vert s-D(E(s))\vert$. $s$ represents a stack of 13 georeferenced images of 0.3 meters in pixel resolution. The images stem from regularly gridded rasters of local spatial statistics over the irregular, three-dimensional LiDAR point cloud. In stage two, the \textit{feature vector} $c=E(s)$ serves as input to a regression deep neural network $R$ to predict an averaged ambient temperatures $t=R(c)$ given by the co-registered Landsat 8 surface temperature product $t$. 
Once the deep neural networks $E$, $D$, and $R$ are trained, we apply a novel inference procedure exploiting backpropagation. The temperature variation $\Delta t$ gets backpropagated through $R$ (denoted by $R^{-1}$) to vary the corresponding feature vector, $c\to c+\Delta c=c+R^{-1}(t+\Delta t)$. In turn, modified LiDAR statistics follow according to:
\begin{align}
s' &=D(c+\Delta c)=D\left(E(s)+R^{-1}\left[R(E(s))+\Delta t\right]\right)\nonumber\\
   &=s'(s,\Delta t)\quad.
   \label{eq:DeepLCZMath}
\end{align}
We note that our approach is generic beyond the application presented here. In the language of \textit{self-supervised learning} (SSL) \cite{balestriero2023cookbook} with recent successes in Earth observation \cite{9875399}, we pretrain a model ($D\circ E$) to generate a (compressed) feature representation $c$ from a remote sensing modality $s$. Subsequently, another co-registered modality $t$ serves to train a \textit{downstream model} ($R$) correlating $c$ to $t$. Thereafter, backpropagation ($R^{-1}$) on frozen models lets us explore the variation $\Delta c$ given variations in $\Delta t$. The procedure allows us to investigate the structure of the feature space modelled by $E$. If there exists an inversion ($E^{-1}\approx D$) to associate $c$ with the input remote sensing modality $s$, the impact of variation $\Delta t$ in the downstream modality on the input modality $s$ helps to explain Earth observation phenomena associated with both modalities, $s$ and $t$.

Given a variation in ambient temperature $\Delta t\in\mathbb{R}$, there exists various choices of feature vector variations $\Delta\vec c\in\mathbb{R}^n$ in $n$-dimensional feature space. $R^{-1}$ is not the \textit{inverse} of $R$! Guided by a \textit{gradient principle}, we label our approach in \Cref{eq:GradFeatureVariationEfficient} ``most effectively'' such that we pose the question: \textit{How to most efficiently vary a geospatial scene's ambient temperature by modifying its urban land use structures represented by LiDAR statistics?}

We assume a gradient vector $\vec{g}=\nabla R\in\mathbb{R}^n$ of downstream model $R$ with components $g_i=\partial R/\partial c_i$ according to feature vector $\vec c$ components $c_i$ such that $\vec g\cdot\vec g=g^2=\sum_jg_j^2$. We define the feature vector variation $\Delta\vec c=\zeta\vec g/g$ parallel to the gradient $\vec g$ scaling it by $\zeta\in\mathbb{R}$ to match a given temperature variation $\Delta t\in\mathbb{R}$:
\begin{equation}
\Delta t=\Delta\vec c\cdot\vec g=\zeta\vec g\cdot\vec g/g=\zeta g\quad.
\end{equation}
Correspondingly, we obtain the model $R$-specific feature vector variations for given temperature variation $\Delta t$:
\begin{equation}\label{eq:GradFeatureVariationEfficient}
\Delta\vec c(R,\Delta t)=\frac{\Delta t}{g^2}\vec g~\text{, in components:}~\Delta c_i = \Delta t \frac{g_i}{\sum_j g_j^2}\quad.
\end{equation}
Indeed: $\Delta\vec c\cdot\vec g=\vec g\cdot\vec g~\Delta t/g^2=\Delta t$. The variation $\Delta\vec c$ parallel to $\vec g$ guarantees by nature of the gradient $\nabla R$ that local, small variations $\Delta c=\sqrt{\Delta\vec c\cdot\Delta\vec c}\ll1$ most strongly, i.e.\ ``most effectively'', modify $t\to t+\Delta t.$
\begin{figure}[t!]
\centering
\hspace{-4ex}\includegraphics[width=.52\textwidth]{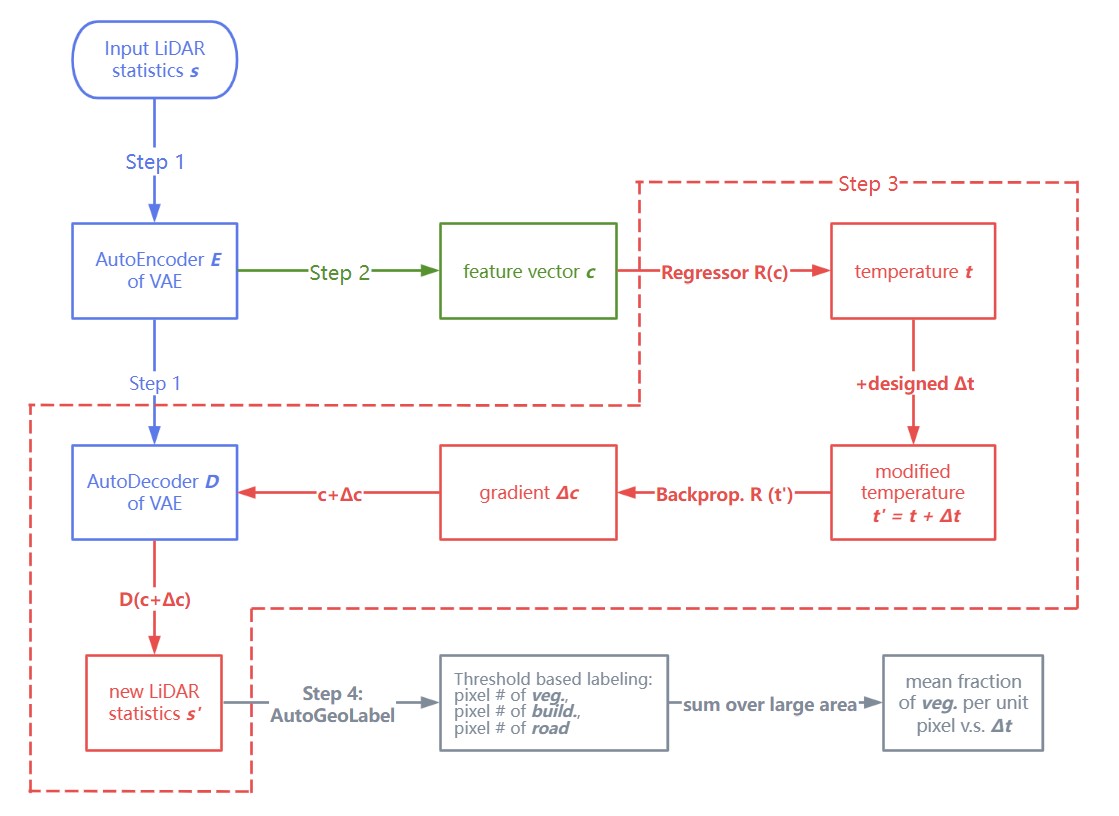}
\caption{\textit{DeepLCZChange} pipeline: Data flow color-coded according to \Cref{fig:architecture}. We color gray the \textit{AutoGeoLabel} post-processing that infers the fraction of vegetation in a scene, cf.\ red data points in \Cref{fig:result}.}
\label{fig:method}
\end{figure}

\Cref{fig:method} depicts the full \textit{DeepLCZChange} processing pipeline---basis to generate our central result, \Cref{fig:result} as further detailed in \Cref{sec:experiment}. We compute the modified LiDAR statistics $s'$ according to \Cref{eq:DeepLCZMath} depending on temperature variation $\Delta t$ and input LiDAR statistics $s$. It allows us to study deviations $\Delta s=\vert s-s'\vert$ depending on $\Delta t$. In particular, by utilizing \textit{AutoGeoLabel} post-processing \cite{albrecht2021autogeolabel} (gray in \Cref{fig:method}), we are able to correlate the fraction of vegetation in a scene with ambient temperature variations.

\section{Experiments}
\label{sec:experiment}

\quad\textbf{Data and area of study.}
We employ the 2017 LiDAR survey of New York City \cite{NYCLidar2017} with focus on the Queens borough to generate raster layers based on LiDAR laser pulse characteristics. We include elevation information (laser light's time-of-flight measurements), the laser pulse return count, and the reflected laser light's intensity in order to generate gridded, spatial statistics \cite{albrecht2021autogeolabel}. As demonstrated in the literature~\cite{albrecht2022monitoring}, such statistics bear signature of human infrastructure such as buildings, vegetation, and traffic networks.

We co-register the post-processed LiDAR statistics and Landsat 8 infrared temperature rasters \cite{malakar2018operational} (pixel resolution approx.\ 100 meters) to conduct our experiments. Specifically, we employ a mid-June snapshot of Landsat 8, band 10, Thermal Infrared Sensor (TIRS 1). We did curate a dataset with 100k samples $(s,t)$ where $s\in\mathbb{R}^{128}\times\mathbb{R}^{128}\times\mathbb{R}^{13}$ and $t\in\mathbb{R}$ covering areas of about 20$\times$20 square meters. The training\slash testing split for our deep learning experiments reads 80k/20k.

\vspace{1ex}
\textbf{Stage 1: feature vector $c$ generation.}
We train a VAE $D\circ E$ \cite{Subramanian2020} to compress\footnote{A $13\cdot(2^6)^2\approx53k$ raw pixels from a 13 channels $128\times128$ square image get reduced to a $2^{10}\approx1k$--dimensional feature vector.} the input LiDAR statistics $c$, cf.\ \Cref{sec:method}. The model was optimized over 100 epochs at learning rate $10^{-3}$. Unlike ordinary RGB images with highly correlated color channels, our input data stacks 13 layers of LiDAR statistics encoding information on elevation and laser pulse reflectance characteristics. Besides the reconstruction loss, a VAE is optimized to bring the distribution of feature vector components $c_i$ close to a Gaussian with mean 0 and standard deviation 1 by virtue of the Kullback-Leibler divergence (KLD). We dynamically adjust the weight $\lambda$ of the KLD relative to the reconstruction loss. $\lambda$ linearly ramps up from $0$ to $10^{-5}$ for the first 50 epochs to stay constant thereafter.

\vspace{1ex}
\textbf{Stage 2: regression network $R$.}
We utilize the compressed feature vector $c\in\mathbb{R}^{1024}$ encoding characteristics of $400m^2$-sized urban scenes to predict its mean surface temperature $t\in\mathbb{R}$ as sensed by Landsat 8. In contrast to directly modelling $t=t(s)$, the VAE encoding guarantees smooth variation of the regenerated scene $s'=D(c)$ on variation of $c$. The three-layer, fully connected regression network $R$ we train by a standard L1-norm loss $\vert R(E(s))-t\vert$. We randomly sample 100 tuples $(t,s)$ from the training set. The errors of the trained model vary in $(-0.8K,+2.1K)$---an acceptable maximum uncertainty of $\sim10\%$ relative to the full range of $20K$ in temperature variation.

\vspace{1ex}
\textbf{Stage 3: backpropagate temperature variation $\Delta t$.}
Once the networks $E$, $D$, and $R$ got trained, their weights $w$ are frozen for the backpropagation of temperature variations $\Delta t$ to the input feature vector $c$, cf. \Cref{eq:DeepLCZMath}. In analogy to the update of weights $w\to w+\eta\partial L/\partial w$ governed by the loss $L$, we update the feature vector components according to the regression model $R$, i.e.\ $c\to c+\zeta\partial R/\partial c$ with the sign of $\zeta$ (as with $\eta$) defining \textit{gradient ascent} ($+$) or \textit{gradient descent} ($-$), respectively. 30 randomly picked locations dominantly sampled from the Queens borough serve as basis to generate about 15k tuples $(s,s')$ over 8 temperature variations $\Delta t/K\in\{\pm1,\pm3,\pm5,\pm10\}$. These data cover a total area of about 3/4 of a square kilometer. They serve as basis for the statistical analysis summarized in \Cref{fig:result}.

\vspace{1ex}
\textbf{Post-Processing: Rule-based vegetation identification through \textit{AutoGeoLabeling}.}
After Stage 3 varied $c=E(s)$ by $R^{-1}$ given $\Delta t$, cf.\ \Cref{eq:DeepLCZMath}, we obtain the modified LiDAR statistics $s'$. Applying \textit{AutoGeoLabel} to $s'$ allows us to determine the fraction of vegetation per unit area $v\in[0,1]$. \textit{AutoGeoLabel} is a near real-time, rule-based labeling framework applicable to high-quality remote sensing information such as LiDAR \cite{albrecht2021autogeolabel}. Though noisy, these easy-to-generate segmentation maps, cf.\ \Cref{fig:three} (right), are sufficient to reveal ground surface changes such as urban forest degradation \cite{albrecht2022monitoring} on variation of $\Delta t$. We average tuples $(\Delta t, v'$) down to 8+1\footnote{\quad$s'=D(E(s))\approx s$ for $\Delta t=0$ where $v'\approx v$} measurements $(\Delta t, \bar v')$.

\section{Results}
Our experimental setup serves as an initial proof-of-concept. The data collection is limited to a single US metropolitan area correlating a single Landsat 8 surface temperature snapshot in summer. Moreover, the noisy nature of the VAE reconstruction and application of \textit{AutoGeoLabel} adds uncertainty. Thus, trends of change in vegetation as observed by our methodology are noisy for individual scenes at the 20 meter scale. We adopt a statistical approach summarizing the overall trend of surface temperature vs.\ vegetation coverage to account for the low signal-to-noise ratio.

\Cref{fig:result} plots the 9 tuples $(\Delta t, \bar v')$ as red data points. We observe: While cooling ($\Delta t<0$) correlates with an increase in vegetation, warming ($\Delta t>0$) lets drop vegetation with increasing $\Delta t$. In order to quantify the relationship, we assume the null hypothesis $H_0$: \textit{An increased fraction of vegetation in the scene does not correlates with a decrease in ambient surface temperature}. We apply a simple Ordinary Least Squares model for linear regression $\bar v'(\Delta t)=a\Delta t + b$ yielding fitting parameters $a=-6\cdot10^{-4}/K$ and $b=.153$. The blue-shaded area in \Cref{fig:result} indicates the confidence interval for the regression coefficients at 95$\%$ confidence level. The coefficient of determination equates to $R^2=.817$. The $p$-value of $a$ reads $.1\%$, i.e.\ for a standard confidence level of $\alpha=.05$, $\bar v'$ and $\Delta t$ are significantly correlated to reject $H_0$.

\section{Conclusion}
\label{sec:conclusion}
When combined with principles of \textit{AutoGeoLabel}, the general concept of \textit{DeepLCZChange} presents a novel deep learning methodology and data pipeline for Earth observation analytics based on remotely sensed data with little need for human interaction. Our initial findings in studying the interplay of urban forests and local ambient surface temperatures in New York City motivate related questions within the scope of urban climate resilience, such as: \textit{How does the interplay of buildings and vegetation affect meteorological quantities such as humidity and temperature?} And based on the previous work \cite{klein2022urban}: \textit{Do correlations of those parameters with urban planning exhibit distinct qualitative trends given the definition of Local Climate Zones?} Insights related will provide guidance to plan urban spaces accounting for climate-resilient solutions.

The novel aspect of the methodology summarized in \Cref{fig:architecture,fig:method} stems from the unsupervised correlation of co-registered geospatial data with the aid of deep learning to uncover the interaction of, e.g., vegetation and surface temperatures. Based on statistical analysis (of historical data) we demonstrate how to qualitatively approach questions related to climate resilience. In the face of climate change, our work hopes to inspire the development of strategies to mitigate issues such as urban heat islands.

\footnotesize
{\centering\subsubsection*{Acknowledgement}}
This work was funded by the Helmholtz Association through the Framework of HelmholtzAI, grant ID: \texttt{ZT-I-PF-5-01} -- Local Unit Munich Unit @Aeronautics, Space and Transport (MASTr). Conrad Albrecht thanks Levente Klein from the IBM TJ Watson Research Center, Yorktown Heights, NY, USA for inspiring discussions on relevant research directions wrt.\ urban forests. We thank Stefan Kesselheim from the Juelich Supercomputing Center for continuous support for conducting our experiments on the JUWELS Booster compute cluster.

{

%\bibliographystyle{IEEEbib}
%\bibliography{refs.bib}
}
\end{document}